# Unlocking Insights Addressing Alcohol Inference Mismatch through Database-Narrative Alignment

*Sudesh Bhagat, Raghupathi Kandiboina, Ibne Farabi Shihab, Skylar Knickerbocker, Neal Hawkins, Anuj Sharma*

## Abstract

Road traffic crashes are a significant global cause of fatalities, emphasizing the urgent need for accurate crash data to enhance prevention strategies and inform policy development. This study addresses the challenge of alcohol inference mismatch (AIM) by employing database narrative alignment to identify AIM in crash data. A framework was developed to improve data quality in crash management systems and reduce the percentage of AIM crashes. Utilizing the BERT model, the analysis of 371,062 crash records from Iowa (2016-2022) revealed 2,767 AIM incidents, resulting in an overall AIM percentage of 24.03%. Statistical tools, including the Probit Logit model, were used to explore the crash characteristics affecting AIM patterns. The findings indicate that alcohol-related fatal crashes and nighttime incidents have a lower percentage of the mismatch, while crashes involving unknown vehicle types and older drivers are more susceptible to mismatch. The geospatial cluster as part of this study can identify the regions which have an increased need for education and training. These insights highlight the necessity for targeted training programs and data management teams to improve the accuracy of crash reporting and support evidence-based policymaking.

## Introduction

Traffic crashes occur worldwide, claiming approximately 1.19 million lives each year and leaving between 20 to 50 million people with non-fatal injuries (Road Traffic Injuries, 2023). Road crashes are the leading cause of death for children and young adults aged 5 to 29 years (Bari et al., 2020). These crashes can happen for various reasons, including environmental, infrastructural, mechanical, medical, and behavioral factors.

For a comprehensive analysis of traffic crashes, it is crucial to record the causal and influencing factors immediately at the crash site. Crashes are primarily documented for legal, medical, and analytical purposes by law enforcement, paramedics, and other officials. This practice ensures the reliability of data, which is essential for accurate estimates and effective policy development. However, many crashes are not documented with complete information, leading to the mismatch of incidents. A mismatched crash is one that lacks essential details regarding factors such as severity, cause, and external conditions. In contrast, misreported or misrepresented crashes involve false information related to the crash factors.

Despite various efforts, data quality can be compromised at several stages, including initial reporting and recording by officers at the scene, data entry into crash databases at data centers, and the subsequent compilation of data for integration with other sources. The accuracy of crash reporting can be affected by factors such as the extent of vehicle damage, the severity of injuries, the willingness of passengers to report, and the discretion exercised by responding officers (Wood et al., 2016). As a result, crashes can be either underreported or misreported. Among all the issues, misinformation regarding alcohol involvement in a crash is particularly concerning, as it can mislead policymakers by providing an inaccurate representation of crash causes. These errors occur because crash reporting systems often follow preset templates that include dropdown menus, checkboxes, and input boxes for descriptive text narratives (Lopez et al., 2022). Furthermore, reporting officers typically receive little to no guidance on how to write the crash narratives (Lopez et al., 2022). Consequently, many reported crashes are inaccurately described, which in turn affects the variables used in crash prediction models (Lopez et al., 2022).

For example, a crash might be reported as non-alcohol-related, even though the narrative states that the "driver smelled of alcohol, but no test was conducted." Conversely, a crash may be labeled as alcohol-related while the narrative mentions that "the passenger was intoxicated, but the driver was not impaired." Therefore, this study focuses on addressing the issue of Alcohol Inference Mismatch (AIM), specifically analyzing crashes where alcohol involvement is not reported despite being a potential contributing factor based on the crash report.

Research indicates that the risk of road crashes significantly increases with a driver's blood alcohol concentration (BAC). A BAC exceeding 0.04 g/dL (grams per deciliter) (Road Traffic Injuries,2023) considerably elevates crash risk, although legal BAC limits in many countries are set at 0.08 g/dL. Several studies have shown that drivers with BAC levels as low as 0.01 to 0.07 g/dL have been involved in fatal crashes. In 2022, reports indicated that 13,524 fatalities in the United States were attributed to drunk driving incidents involving BAC levels above 0.08 g/dL (Drunk Driving, Statistics and Resources, NHTSA, 2022). However, the actual number of alcohol-related crashes, particularly non-fatal ones, is likely higher due to mismatch and misreporting. While many studies have investigated the extent of this mismatch in alcohol-related crashes, relatively few have proposed comprehensive solutions. In response, the current study aims to develop a framework that can be integrated into crash data pipelines to detect crashes involving alcohol mismatched as non-impaired incidents. This study specifically investigates cases where crash narratives contain alcohol-related terminology but have been misclassified.

To achieve this, an advanced Natural Language Processing (NLP) model, specifically the BERT Large Uncased model, has been utilized to analyze extensive crash narrative data from the Iowa Department of Transportation (DoT) crash database. Additionally, statistical tools have been applied to identify factors that contribute to crash mismatch. It is anticipated that understanding these factors will lead to the development of targeted recommendations aimed at improving the accuracy of crash reporting practices.

The main objectives of this study are:
1. To build an NLP model that holistically analyzes crash narratives to infer alcohol involvement by capturing contextual, linguistic, and semantic patterns.
2. To apply the developed model to uncover patterns in reference to alcohol in crash reports that have not been marked in the database, enhancing training and accuracy in reporting.

## Literature:

The mismatch of crashes has been a well-known issue all over the world for decades (Alsop and Langley, 2001; Amoros et al., 2006; Dandona et al., 2008; Salifu and Teresa, 2009; Watson et al., 2015). Despite the developments in technology and advancements in crash reporting systems, this issue persists. While factors such as age, gender, and road geometry can be collected later and updated in the system, critical factors such as driving under influence, distraction, and protection need to be collected immediately after the crash at the crash site. These factors are sometimes difficult to identify or estimate, especially when driving under the influence of alcohol needs to be tested within six hours from the incident (MedlinePlus, 2007).

Research indicates that fatal crashes are more likely to be reported accurately due to their clear definitions and severe consequences. However, non-fatal crashes frequently remain underreported due to ambiguities in classification (Amoros et al., 2006). For instance, a study conducted in Denmark found that while 97% of fatal crashes were reported by law enforcement, only 47% of non-fatal crashes and 10% of crashes involving cyclists were documented when comparing hospital and police records (Elvik Rune and Mysen Borger Anne, 1999). This mismatch typically involved crashes that were never reported to any enforcement agency rather than omissions of specific crash factors. The alcohol inference mismatch rate across Europe was estimated to range from 25% to 57% (Elvik Rune and Mysen Borger Anne, 1999). Crashes involving children under 18 years old, females, severe injuries, pedestrians, seatbelt or helmet use, high-speed roads, multilane roads, and those occurring during morning peak hours and weekdays were more likely to be reported (Janstrup et al., 2016). Similarly, another study found that while only 0-9% of fatal crashes showed AIM, non-injury crashes showed AIM rates of 46-65%, while injury crashes resulted in AIM rates between 7% and 80% (Wood et al., 2016).

While much of the literature focuses on general crash mismatch, some studies have specifically examined AIM. Definitions of alcohol-related crashes vary significantly across regions, contributing to discrepancies in reported data. For example, the European SafetyNet project (2008) defined alcohol-related crashes as "any death occurring within 30 days as a result of a fatal road crash in which any active participant was found with a BAC level above the legal limit" (Salifu and Teresa, 2009). In Europe, the legal BAC limit ranges from 0.05 to 0.08 g/dL, while in the United States, alcohol-impaired driving crashes are defined as "crashes that involve at least one driver or motorcycle operator with a BAC of 0.08 grams per deciliter (g/dL) or higher" (Bozak David J, 1996; NHTSA, 2022). Pedestrians and cyclists are generally excluded from these definitions, as most jurisdictions lack enforceable BAC limits for them.

Testing for alcohol-related crashes remains inconsistent due to administrative and procedural challenges. In many countries, governments have yet to establish comprehensive protocols for BAC testing following road traffic crashes (Alsop and Langley, 2001). Even where such procedures exist, bureaucratic hurdles and extensive paperwork hinder effective testing, contributing to the mismatch. BAC-related reports may be misplaced, and law enforcement officers may misjudge alcohol levels, further exacerbating the issue. Additionally, some jurisdictions prohibit post-mortem BAC analysis following fatal crashes, complicating accurate reporting (Alsop and Langley, 2001).

Research has highlighted the severity of AIM in crashes. A US-based study analyzing police and hospital data from 2006 to 2008 found significant disparities in alcohol involvement reporting. Police documented alcohol involvement in only 44% of cases identified by hospitals, while hospitals reported alcohol involvement in just 33% of cases documented by police (Miller et al., 2012). Improved communication and data-sharing between these sources were recommended to reduce the mismatch. Another study explored using large language models to detect AIM in crashes. Researchers employed three models—ChatGPT, LLaMA-2, and Flan-UL2—through prompt engineering. They processed 500 crash narratives using three independent human annotators and conducted follow-up analyses when annotators reached different conclusions. The combined annotations were compared with initially reported data to detect the mismatch (Arteaga and Park, 2025). In a related study, researchers examined how varying definitions, legislation, and reporting procedures influenced the documentation of alcohol-related crashes (Vissers et al., 2018). They developed a questionnaire covering four categories: drink-driving legislation, definitions of alcohol-related casualties, recording methods, and the quality of official statistics. The findings revealed inconsistencies in data collection methods, emphasizing the need for standardized definitions and improved data reliability (Vissers et al., 2018).

Additional research investigated discrepancies in BAC reporting on death certificates across the US by comparing national death data with the Fatality Analysis Reporting System (FARS). The analysis used a "reporting ratio" to measure the extent of alcohol-involvement reporting on death certificates relative to FARS records, revealing significant mismatch across states (Castle et al., 2014). A study focused on non-fatal alcohol-related crashes in the US compared crash reporting accuracy between police and hospitals. Police-related mismatch was found to be 44%, while hospital-related mismatch stood at 33% (Miller et al., 2012). Researchers used a Bayesian-based capture-recapture model to estimate the true extent of mismatch.

Research has also highlighted that law enforcement officers frequently fail to identify BAC levels accurately during crash investigations. One study found that officers reported only 47% of alcohol-related crashes where drivers had consumed alcohol (Blincoe et al., 2023). Reporting rates increased with higher BAC levels, with only 18% of crashes involving BACs between 0.01 and 0.09 being reported, compared to 48.9% of crashes involving BACs above 0.1(Arteaga and Park, 2025). This trend was supported by another study showing that 71% of crashes involving drivers with high BACs were correctly identified (Soderstrom et al., 1990). However, these studies are over a decade old. More recent research indicates that law enforcement officers report alcohol involvement more consistently in less severe crashes than in fatal crashes, possibly reflecting improvements in road safety measures implemented by local and state governments (Blincoe et al., 2023).

Upon reviewing existing studies, a significant research gap emerges concerning the identification and correction of underreported alcohol-related crashes within crash databases. While previous research has extensively explored the causes and extent of alcohol inference mismatch, few studies have proposed data-driven solutions for correcting this issue. The present study seeks to bridge this gap by leveraging automated data-processing techniques to identify crashes with AIM within crash narratives. The goal is to minimize the mismatch through improved post-processing of crash reports, enabling more accurate data-driven policy development and road safety interventions. In addition, the identification of counties with a high mismatch percentage will aid in improving law enforcement through policy changes training.

## Data Description:

The Iowa Department of Transportation (DoT) maintains a comprehensive crash database that includes detailed information related to crashes, such as personal details, vehicle information, weather conditions, road characteristics, medical data, and other relevant factors. This information is primarily reported by law enforcement agencies, highway patrols, and paramedics, following the recommendations outlined in the latest Strategic Highway Safety Plan (SHSP) for Iowa DoT (2019) (SHSP, 2024). The investigating law enforcement officer completes the "Investigating Officer's Report of Motor Vehicle Accident" form developed by Iowa DoT (Crash Reporting Guide for Investigating Officers, 2024). These forms are submitted electronically through the Traffic and Criminal Software (TRaCS), which is accessible to all law enforcement agencies in Iowa.

In addition to structured data, these forms include sections for diagrams and crash narratives (Crash Reporting Guide for Investigating Officers, 2024). According to the Model Minimum Uniform Crash Criteria (MMUCC) guidelines issued by the National Highway Traffic Safety Administration (NHTSA), law enforcement officers are expected to exercise their judgment in determining whether alcohol involvement is suspected in a crash (Highway Traffic Safety Administration, 2024). This reliance on subjective judgment can introduce AIM in crash reporting due to various factors. For instance, if an officer suspects alcohol involvement but a BAC test is not conducted, the crash may be reported as non-alcohol-related, despite the officer's suspicions. Furthermore, crash reporting systems may prioritize structured data fields over narrative descriptions, which could lead to overlooking officers' qualitative judgments and contributing to AIM.

The present study utilized the crash database for a detailed analysis. The crash database contains essential information, including crash narratives, which are critical for this research. Researchers were granted authorized access to this confidential data after completing relevant training on proper data handling procedures. The analysis covers crash data from 2016 to 2022, encompassing all types of crashes. This data is extracted from three interrelated tables: driver information, crash information, and crash narratives, all linked by the unique identifier "CRASH_KEY" assigned to each crash.

The dataset includes a total of 371,062 crashes categorized into five severity levels defined by the KABCO scale (KABCO Injury Classification Scale and Definitions, 2014). Table 2 illustrates the number of crashes by severity level, indicating that most crashes are classified as "property damage only," with counts decreasing as severity increases. The final merged dataset features various variables listed in Table 1, with "ALCOHOL_REL" and "CRASH_NARRATION" being crucial for identifying crashes with potential alcohol inference mismatches. Yearly crash counts, shown in Table 2, indicate consistent totals across the years, except for 2020 and 2022, likely due to restrictions related to the COVID-19 pandemic.

**Table 1: Crash Data**

| Attributes | Description | Categories |
|---|---|---|
| CRASH_KEY | Unique Identifier | Case Number |
| DriverGen | Gender of Injured Person | Male or Female |
| DriverAge | Age of Injured Person | 0-25 Years, 26-64 Year', Over 65 Years |
| COUNTY | Location | Name of the Counties |
| WEATHER | Weather Condition | Clear, Cloudy, Rain, Snow, Fog, smoke, smog and Severe winds |
| LIGHT | Light Condition | Dark, Daylight and Dusk |
| ROADTYPE | Roadtype | Intersection and Non-Intersection |
| CSEVERITY | Severity level | Property Damage Only, Minor Injury, Possible/Unknown, Major Injury and Fatal |
| CRASH_YEAR | Crash Year | 2016,2017,2018,2019,2020,2021,2022 |
| CRASH_DATE | Crash Date | Date of the Crash |
| DRIVERDIST | Driver Distraction | Driver Distracted and Not Distracted |
| WZ_RELATED | WorkZone Related | WorkZone and Non-Workzone |
| SPEED_LIMIT | Posted Speed | <45 Mph, 45-55 Mph and over 55 Mph |
| ALCOHOL_REL | Whether alcohol related | Alcohol and Non-Alcohol |

| CRASH_NARRATION | Description of police report | Processed Text |
|---|---|---|
| Vehicle type | Type of vehicle | heavy_trucks,motorcycles,other_vehicles and Cars |
| Road user | Type of Road User | Pedestrians and bicyclists |
| Unprotected person | Seat Belt | Yes and No |
| Season | Different Seasons | Winter ,Spring ,Summer,and Fall |
| Functional Class Type | Roadway Classification | Major Roads, Arterial Roads ,Collector Roads and Local Roads |
| RURALURBAN | Geographical Demographic | Urban or Rural |
| AADT | Annual Average Daily Traffic | Traffic Volume |

The 'ALCOHOL_REL' column in the final merged dataset is a descriptive variable indicating the involvement of alcohol as reported by the officer. This is a binary variable categorized as 'Alcohol-related' and 'Non-Alcohol-related,' as shown in Table 1. Table 2 illustrates the distribution of these categories across different severity levels. The percentage of 'Alcohol-related' crashes increases with crash severity, with nearly one-quarter of fatal crashes recorded in the database involving alcohol.

**Table 2: Reported Alcohol Crashes**

| Years | Property Damage only | | Possible/Unkown | | Minor Injury | | Major Injury | | Fatal Crashes | | Total | |
|---|---|---|---|---|---|---|---|---|---|---|---|---|
| | Alcohol | Non-Alcohol | Alcohol | Non-Alcohol | Alcohol | Non-Alcohol | Alcohol | Non-Alcohol | Alcohol | Non-Alcohol | Alcohol | Non-Alcohol |
| 2016 | 1077 | 38557 | 324 | 8965 | 364 | 4535 | 196 | 990 | 89 | 236 | 2050 | 53281 |
| 2017 | 1113 | 38421 | 325 | 9027 | 397 | 4531 | 165 | 1039 | 74 | 193 | 2074 | 53211 |
| 2018 | 975 | 40140 | 266 | 8678 | 357 | 4479 | 151 | 896 | 69 | 183 | 1818 | 54376 |
| 2019 | 1031 | 41366 | 265 | 8831 | 365 | 4621 | 139 | 945 | 75 | 205 | 1875 | 55768 |
| 2020 | 1023 | 32806 | 274 | 7230 | 387 | 3935 | 169 | 874 | 57 | 206 | 1908 | 50329 |
| 2021 | 1173 | 37500 | 295 | 8320 | 376 | 4536 | 180 | 989 | 63 | 228 | 2087 | 51573 |
| 2022 | 1856 | 31741 | 1047 | 7040 | 367 | 4050 | 138 | 903 | 62 | 196 | 3470 | 43930 |
| % alcohol crashes | 3.06 | | 4.59 | | 7.85 | | 14.64 | | 25.26 | | 30.14 | |

All variables in the database have defined categories, except for the crash narration. The crash narration provides a detailed description of the crash situation based on field observations, which include statements from law enforcement officers, eyewitnesses, and drivers. Although there is no fixed format or word limit for these narrations, the naming conventions for the vehicles and individuals involved remain consistent. The narrations can vary significantly in length, ranging from as few as ten words to over a thousand words. Due to the lack of a standard format, Natural

Language Processing (NLP) tools are used to structure these narrations for further analysis. Once processed, the crash narrations are integrated with the original crash dataset using the "CRASH_KEY" to enable comprehensive analysis.

## Methodology

The study has two primary objectives. The first objective is to develop a method for identifying AIM crashes based on the available crash database. The second objective is to assess the impact of various crash attributes on the likelihood of AIM crashes through statistical modeling, enabling the identification of regions where law enforcement can be improved.

To achieve the first objective, the study focuses on identifying crashes that are linked to alcohol involvement based on evidence found in the crash narratives, even if they are reported as 'Non-Alcohol-related' in the "ALCOHOL_REL" field of the dataset. This process involves analyzing crash narratives for keywords or statements that may indicate alcohol involvement. Once the mismatched alcohol crashes are identified, the AIM percentage is calculated as the ratio of the number of AIM crashes to the total number of alcoholic crashes in a given region during a specific time period.

Given the substantial volume of data, manually reviewing these narratives for terms related to alcohol is impractical. Therefore, an NLP model is trained on a prepared sample of crashes to classify the narratives as "Alcoholic" or "Non-Alcoholic." Importantly, the study ensures the protection of Personally Identifiable Information (PII), such as names, phone numbers, vehicle numbers, and other personal details reported in the narratives, for legal and medical purposes. The spaCy package (Presidio, 2016; ArcGIS Pro) was utilized for this purpose. Finally, crashes are classified for 'Mismatch Category' as "AIM Crash" or "Non-AIM Crash" based on criteria outlined in Table 4. The AIM percentage is calculated by dividing the number of identified AIM crashes by the total number of crashes, which includes both Non-AIM Crashes and those identified as AIM. This result is then multiplied by 100 to express the AIM as a percentage.

**Table 4: Crash 'Mismatch Category'**

| Mismatch Category | Original label | Predicted label |
|---|---|---|
| AIM Crash | Non-Alcohol | Alcoholic |
| Non-AIM Crash | Alcohol | Non-Alcoholic |

To achieve the second objective, the labeled 'Mismatch Category' data from the previous step is used to develop statistical models, such as a probit logit model at the crash level, to identify the factors influencing AIM crashes based on selected attributes listed in Tables 1 and 3. The detailed methodology for this two-stage analysis is illustrated in Figure 1.

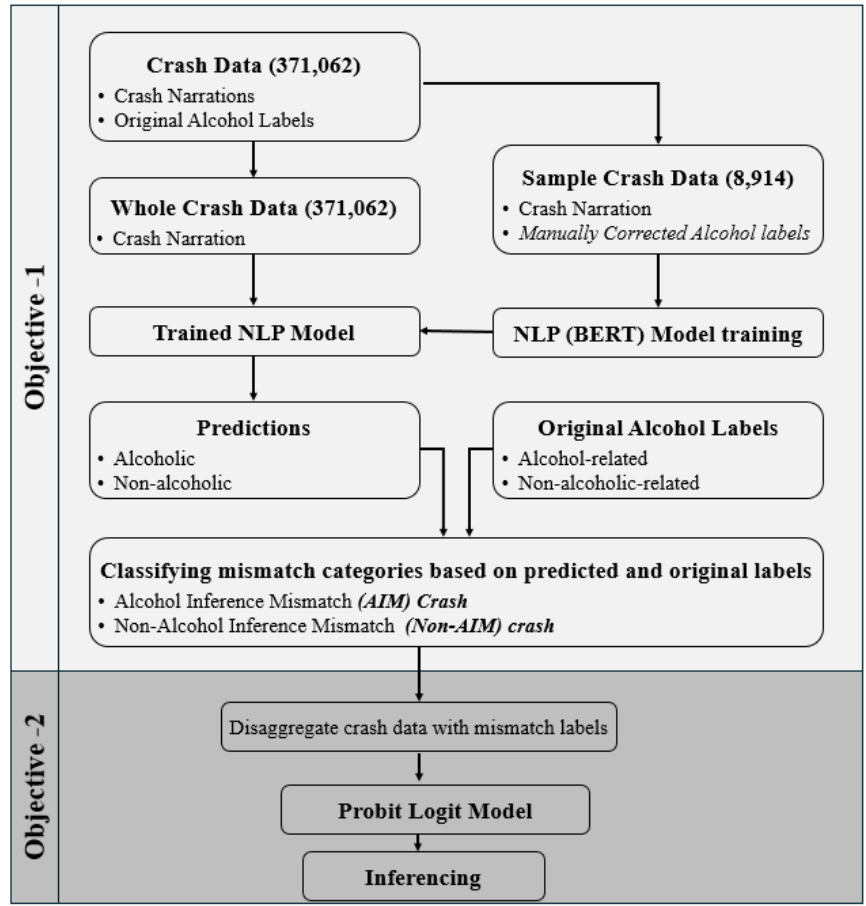

**Figure 1: Methodology**

## Analysis

### *Objective 1: Framework for identifying AIM crashes using BERT Text Classification*
*BERT Text Classification*

The study employed the BERT Large Uncased model, a bidirectional, unsupervised language representation model known for its capability to handle complex linguistic structures (Janstrup et.al., 2016). The framework for achieving this objective is illustrated in Figure 1. BERT Large Uncased was selected due to its deep-layered architecture and substantial parameter capacity, enabling a more nuanced understanding of sentence semantics compared to simpler models. The crash narrative data underwent extensive pre-processing to ensure optimal model performance. Personally, Identifiable Information (PII), punctuation, and numerical values were removed to maintain data privacy and standardize the text input. Lemmatization was applied to reduce words to their base forms, followed by splitting the dataset into training and validation subsets.

To improve feature extraction from the crash narratives, the study applied Term Frequency (TF) and Inverse Document Frequency (IDF). TF counts the frequency of a term within a document, while IDF measures how unique or rare the term is across all documents (Geetha and Karthika Renuka, 2021). The product of TF and IDF yielded weighted scores, with higher values indicating rarer and potentially more informative terms. The BERT Large Uncased model was trained on these weighted features to classify crashes based on alcohol involvement. Model evaluation metrics such as precision, recall, and F1-score were employed to assess the model's performance. Specific keywords considered indicative of alcohol involvement included terms like "influence," "consuming," "odor," and "impaired". Automating crash narrative analysis through the BERT model significantly reduced the manual effort required for identifying underreported alcohol-related crashes.

*Preparing Training Dataset and Training the Model*

A stratified sample of 8,914 crashes was carefully selected after manually correcting the "ALCOHOL_REL" field by reviewing individual narratives. This correction was necessary due to various errors that can occur in reports, as discussed in the introduction. Each narrative labeled as "Alcohol-related" in the "ALCOHOL_REL" attribute was individually reviewed to verify the presence of alcohol involvement. Similarly, narratives labeled as "Non-Alcohol-related" were examined to ensure they did not mention alcohol. This meticulous validation process helped maintain the integrity and reliability of the dataset before training and testing the model.

The validated sample was divided into training and testing sets in an 80:20 ratio. The BERT Large Uncased model was trained on the training set and evaluated on the test set to predict the "ALCOHOL_REL" attribute. The model achieved accuracy rates of 97% for the training set and 98% for the testing set, as shown in Table 5. The predictions and original labels are presented in the confusion matrices for both the testing and training sets in Figure 2. These results demonstrate the effectiveness of the BERT model in classifying crashes as either alcohol-related or non-alcohol-related. The high accuracy levels of the BERT models align with findings from other studies, such as one (Devlin et al., 2019) in which BERT attained an F1 score of 93.2 on the SQuAD v1.1 dataset (Stanford Question Answering Dataset developed by Stanford University) and 94.9% accuracy on the MNLI (Multi-Genre Natural Language Inference) dataset. Overall, the results confirm that the BERT Large Uncased model effectively distinguishes between alcohol-related and non-alcohol-related crashes based on the content of the narratives.

**Table 5: Training and Testing Accuracy**

| Data | Category | Precision | Recall | F1-Score | Support |
|------|----------|-----------|--------|----------|---------|
| **Training** | Alcohol | 0.99 | 0.96 | 0.97 | 3559 |
| | No Alcohol | 0.96 | 0.99 | 0.97 | 3142 |
| | **Accuracy** | | | **0.97** | |
| **Testing** | Alcohol | 0.99 | 0.97 | 0.98 | 898 |
| | No Alcohol | 0.97 | 0.99 | 0.98 | 902 |
| | **Accuracy** | | | **0.98** | |

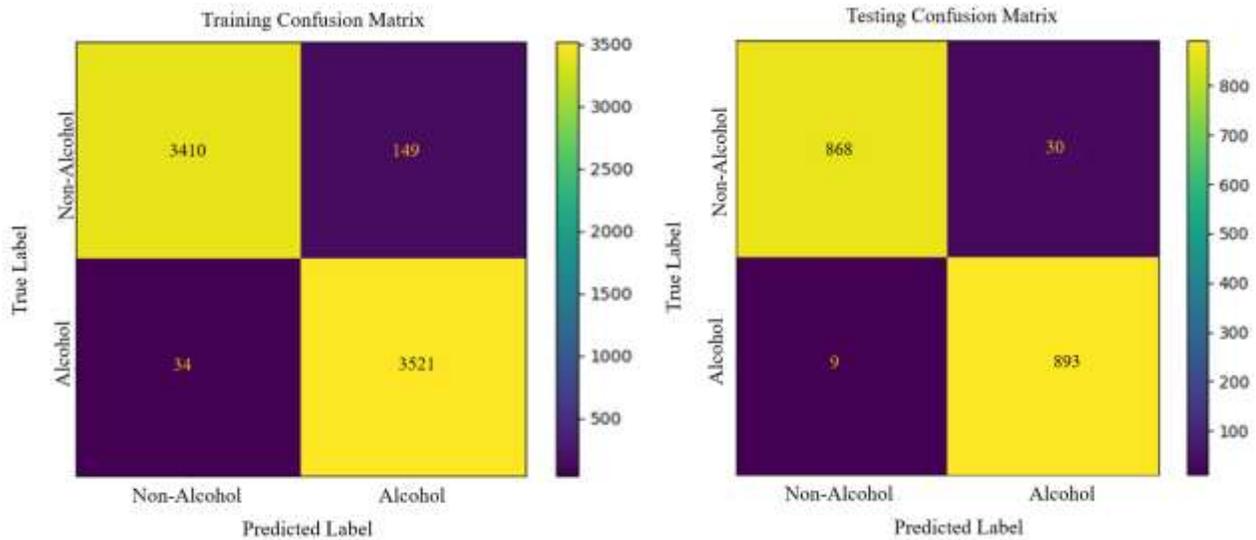

**Figure 2: Training and Testing Confusion Matrix**

*Analysis on the whole data and AIM Percentage Estimation*

Given the satisfactory performance of the trained NLP model, it was applied to the entire dataset covering the years 2016 to 2022, which AIM includes 371,062 crashes. The primary objective of this study is to identify AIM crashes. Crashes predicted as 'Non-Alcoholic' by the model are not relevant to this study, as they are assumed to be accurate. Therefore, we filtered out only those crashes predicted as "Alcoholic" by the model. These filtered crashes were then categorized as either AIM or Non-AIM, as shown in Table 4.

The filtered dataset comprised approximately 11,517 records, with 2,767 categorized as "AIM crashes" and 8,750 as "Non-AIM crashes". Consequently, the overall AIM percentage, or AIM rate, was estimated to be 24.03%. The variation in AIM percentage over the selected years is illustrated in Figure 3. The fluctuation in the AIM rate across these years was not significant, remaining between 22% and 25%. The lowest AIM rate was observed in 2022, despite that year's total alcohol-related crashes being the highest. Furthermore, an increasing trend was noted in the years leading up to the pandemic, followed by a decrease during the pandemic years.

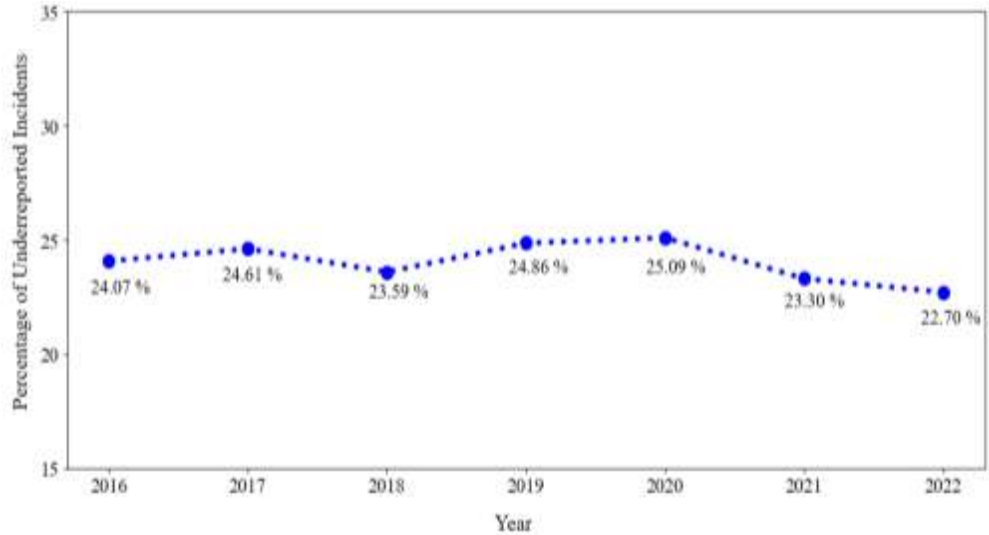

**Figure 3: Variation of AIM Percentage from 2016-2022**

In contrast, the variation in AIM rate among different severity levels was more pronounced. Fatal crashes had the lowest AIM rate at 16.79%, while crashes with unknown injuries had the highest AIM rate at 32.48%, as depicted in Figure 4. This aligns with the logical expectation that fatal crash reports are more meticulously prepared, whereas uncertainties are more likely to occur in crashes of unknown severity.

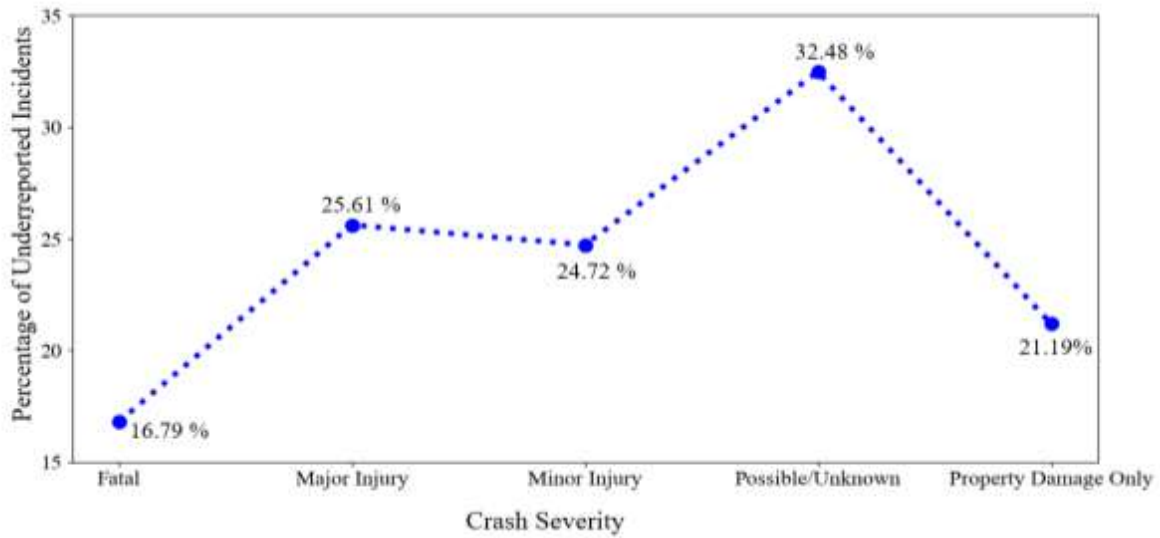

**Figure 4: AIM Percentage Based on Severity-Type**

***Objective:2 Assessing the impact of crash attributes on AIM to Inform Data-Driven Enforcement Strategies***

*Estimating the Spatial Correlation at county level*

Once the "Mismatch Category" was assigned to each crash using the BERT model, the AIM percentage was calculated for each county in Iowa to assess spatial autocorrelation. This analysis examines the existence of spatial clusters among Iowa's 99 counties, allowing for targeted improvements in crash reporting methods within these clusters. Spatial patterns can also influence statistical model estimates, potentially leading to incorrect inferences about the fitted variables. Therefore, it is essential to account for spatial autocorrelation, if present, before applying any statistical models to the selected variables.

To examine whether the mismatch is spatially clustered or randomly distributed, Local Indicators of Spatial Association (LISA) specifically Local Moran's I was applied to the mismatch rate at the county level. The analysis revealed a mix of statistically significant clusters and spatial outliers. Notably, counties such as Dallas and Greene were identified as High-High clusters, indicating high mismatch surrounded by similarly high values. In contrast, counties like Audubon and Boone formed Low-Low clusters, where both the county and its neighbors exhibited consistently low mismatch.

Several counties, including Lyon, Emmet, and Jones, appeared as spatial outliers, identified as Low-High clusters, indicating that their mismatch levels differed significantly from those of neighboring counties. Similarly, Osceola, Obrien and Dickinson were identified as High-Low clusters. These results, shown in Table 6 and Figure 5, suggest that while spatial clustering exists in certain areas, the overall spatial pattern of mismatch is not uniform across the state. Although this pattern may be correlated with the mismatch, it may also be due to the fact that the law enforcement officer may have left the report blank. Additionally, high mismatch rates did not necessarily align with counties reporting the highest number of crashes, suggesting that some counties with relatively few crashes may still exhibit disproportionately high mismatch. A probit logistic regression model was used to identify the factors that significantly impacted the mismatch. Such factors also presented the areas with the potential for increased law enforcement.

**Table 6: Local Moran's I and P-value**

| CountyName | LISA_cluster | P_values | Local Morans I |
|------------|--------------|----------|----------------|
| Lyon | Low-High | 0.00 | 0.00 |
| Dallas | High-High | 0.00 | 0.32 |
| Emmet | Low-High | 0.02 | -1.46 |
| Greene | High-High | 0.02 | 1.43 |
| Audubon | Low-Low | 0.03 | -1.62 |
| Jones | Low-High | 0.04 | -0.87 |
| Osceola | High-Low | 0.04 | 1.05 |
| Boone | Low-Low | 0.04 | -0.27 |
| Obrien | High-Low | 0.04 | 0.84 |
| Dickinson | High-Low | 0.04 | 0.54 |

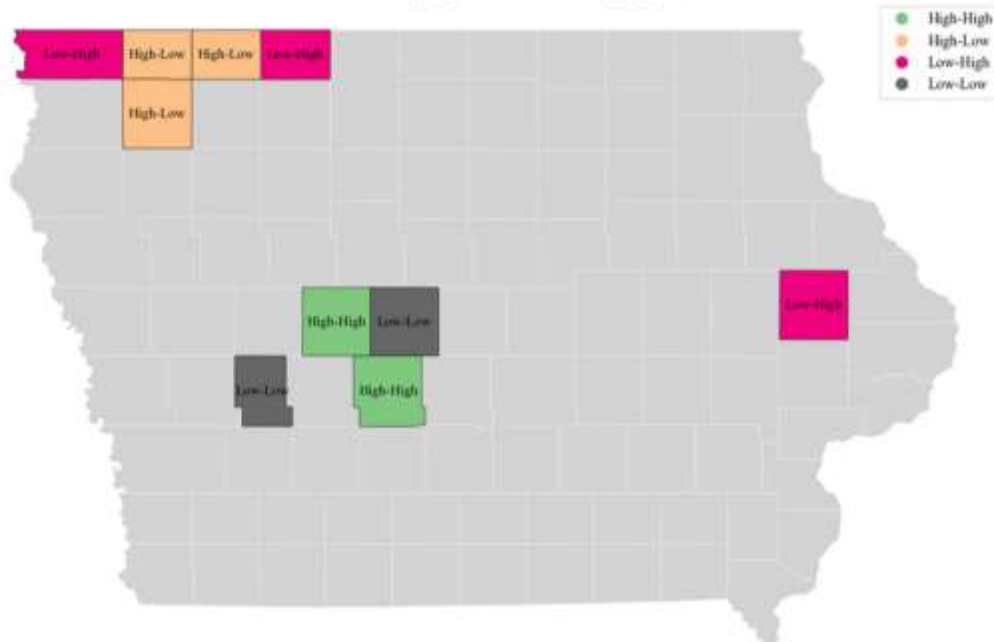

**Figure 5: LISA Cluster Map**

*Regression modeling of predicted "Mismatch Category" of each crash using Probit Logit*

This model is developed at the crash level, where the outcome variable is the "Mismatch Category," classified as either an AIM Crash or Non-AIM Crash. The labeled data for the "Mismatch Category" for each crash, along with other relevant crash variables outlined in Table 1, were used to build the Probit Logit model. The Probit Logit model is particularly suitable for binary outcome variables, making it an ideal choice for this analysis where the dependent variable is dichotomous, classified as either "AIM Crash" or "Non-AIM Crash." Additionally, this model offers latent variable interpretation, which is essential for capturing potential misclassification of crashes as non-alcohol-related, especially when underlying false indicators in the crash narratives are present.

A balanced sample, consisting of 3,895 samples for each category of the "Mismatch Category," was used for the analysis. This sample contained the variables listed in Table 7, all of which are categorical variables and numeric, along with their range.

**Table 7: Variable Categories and Range**

| Variable | Categories | Range |
|---|---|---|
| DriveDistracted | No | 6087 |
| | Yes | 1703 |
| Crash_Severity | Property Damage Only | 3934 |
| | Fatal | 801 |
| | Possible/Unknown Injury | 1531 |
| | Minor Injury | 1524 |
| Weather_Condition | Clear | 5820 |
| | Cloudy | 1150 |
| | Rain | 550 |
| | Snow | 195 |
| | Fog, smoke, smog | 59 |
| | Severe winds | 16 |

| | Daylight, | 3035 |
|---|---|---|
| LightCondition | Dark | 4452 |
| | Dusk | 303 |
| DRIVERAGE | Age_25_to_64_years | 4897 |
| | Age_65_years_and_above | 832 |
| | Age_15_to_24_years | 1767 |
| SPEEDLIMIT | 25-55 Mph | 6833 |
| | over 55 Mph | 756 |
| | <25 Mph | 201 |
| RURALURBAN | Urban | 4681 |
| | Rural | 3109 |
| INJUREDGEN | Male | 5454 |
| | Female | 2336 |
| unprotected_persons | Yes | 778 |
| | No | 7012 |
| Roadtype | Non-Intersection | 5839 |
| | Intersection | 1951 |
| work_zones | No | 7696 |
| | Yes | 94 |
| vehicle_type | car | 7268 |
| | heavy_trucks | 182 |
| | other_vehicles | 340 |
| FEDERAL_FUNCTIONAL_CLASS | MajorRoad | 1765 |
| | ArterialRoads | 1808 |
| | CollectorRoads | 1632 |
| | LocalRoads | 2585 |
| season | Winter | 1710 |
| | Spring | 1905 |
| | Summer | 2089 |
| | Fall | 2086 |
| Road_user | None | 7562 |
| | pedestrians | 166 |
| | bicyclists | 62 |
| Log_AADT | -4.61 to 11.75 | |

A total of 7,790 samples were used to fit a probit logit model with a random intercept for counties, iterated across 20 runs. The best-fitting model—determined by the lowest AIC and BIC values—was selected to explore variability in alcohol inference mismatch across counties. **Figure 6** presents results filtered to include only those counties identified as spatial anomalies through LISA clustering.

To account for unobserved contextual differences across counties—such as enforcement intensity, reporting practices, or crash documentation protocols—county-level variation using a random intercept for each county was explicitly modeled. The estimated variance of the random intercept was 0.052, with a standard deviation of 0.228, indicating moderate variability in the baseline likelihood of alcohol mismatch across counties, as shown in **Table 8**.

*Bhagat et al.*

As illustrated in **Figure 6**, several counties exhibit notable deviations from the overall mean. For example, Emmet and Jones counties have positive random intercepts and are classified as Low-High clusters in the LISA analysis, suggesting that their baseline probability of alcohol mismatch is higher than expected given their surroundings. Conversely, Obrien, Dickinson, and Boone counties exhibit negative random intercepts and are classified as High-Low clusters, indicating potential underperformance relative to neighboring counties. These spatial anomalies underscore the importance of integrating mixed-effects modeling with spatial diagnostics to identify counties where observed reporting patterns diverge meaningfully from model-based expectations.

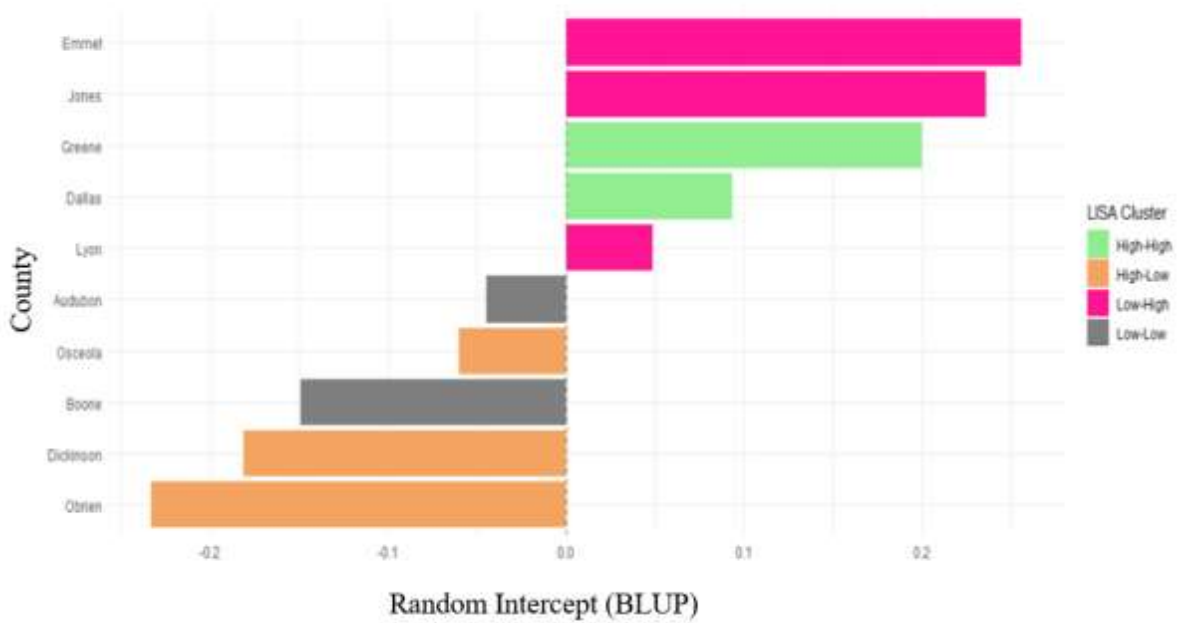

**Figure 6:** Random Effect County Anomalies

Table 8, highlights the statistical significance of various crash characteristics in predicting the likelihood of mismatched alcohol-related crashes.

**Table 8: Model with the Most Significant Variables**

| Random effects: | | | | |
|---|---|---|---|---|
| **Groups** | **Name** | **Variance** | **Std. Dev** | |
| County | Intercept | 0.05205 | 0.2281 | |
| Fixed effects: | | | | |
| **Variable Name** | **Estimate** | **Std.Error** | **Z Value** | **Pr(>\|z\|)** |
| (Intercept) | -1.049 | 0.094 | -11.111 | 0.000 |
| vehicle_typeheavy_trucks | 0.329 | 0.074 | 4.419 | 0.000 |
| vehicle_typeother_vehicles | 0.281 | 0.059 | 4.772 | 0.000 |
| Injured Gender -Male | -0.063 | 0.024 | -2.586 | 0.010 |
| Location-Urban | 0.062 | 0.026 | 2.366 | 0.018 |
| RoadtypeNon-Intersection | 0.033 | 0.026 | 1.266 | 0.206 |
| DRIVERAGEAge_15_to_24_years | 0.198 | 0.027 | 7.235 | 0.000 |
| DRIVERAGEAge_65_years_and_above | 0.620 | 0.038 | 16.243 | 0.000 |
| SpeedLimit <25 Mph | -0.105 | 0.071 | -1.473 | 0.141 |
| SpeedLimit over 55 Mph | -0.020 | 0.082 | -0.246 | 0.806 |
| Road_user-bicyclists | 1.092 | 0.146 | 7.495 | 0.000 |
| Road_user-Pedestrians | 1.067 | 0.088 | 12.092 | 0.000 |
| LightCondition-Daylight | 0.507 | 0.023 | 21.658 | 0.000 |
| LightCondition-Dusk | 0.291 | 0.057 | 5.094 | 0.000 |
| Crash_Severity-MinorInjury | 0.042 | 0.044 | 0.944 | 0.345 |
| Crash_Severity-Possible/UnknownInjury | 0.256 | 0.046 | 5.524 | 0.000 |
| Crash_Severity-Property Damage Only | -0.018 | 0.043 | -0.411 | 0.681 |
| Log_AADT_scaled | -0.024 | 0.015 | -1.545 | 0.122 |
| Functional Class-CollectorRoads | -0.018 | 0.035 | -0.520 | 0.603 |
| Functional_Class- LocalRoads | -0.073 | 0.036 | -2.048 | 0.041 |
| Functional_Class-MajorRoad | -0.013 | 0.036 | -0.377 | 0.706 |
| Unprotected_Persons-Yes | -0.242 | 0.042 | -5.716 | 0.000 |

Crashes involving heavy trucks and other vehicles were significantly more likely to be misreported compared to those involving cars, as observed by the strong positive associations in the model (p < 0.001). Additionally, crashes involving younger drivers (15 to 24 years) and older drivers (65 years and above) also had a higher chance of being classified as AIM crashes. Speed limits above 55 mph and below 25 mph did not have a statistically significant impact on the likelihood of alcohol inference mismatches, suggesting that speed alone does not influence mismatch after accounting for other variables. Crashes involving bicyclists and pedestrians showed much higher log odds of being classified as AIM crashes (p < 0.001) compared to vehicle crashes, indicating potential inconsistencies in reporting incidents involving vulnerable road users. Interestingly, the presence of unprotected persons reduced the likelihood of alcohol inference mismatch (p < 0.001), suggesting that these crashes are more accurately documented.

Moreover, light conditions played a significant role, with daylight and dusk conditions increasing the likelihood of AIM (p < 0.001). In contrast, crashes occurring at night were less likely to show alcohol inference mismatches, possibly due to stricter enforcement and heightened awareness among reporting officials during late hours. Crash severity was also a crucial determinant. Crashes with possible or unknown injuries demonstrated significantly higher log odds of alcohol inference mismatch compared to fatal crashes, which served as the reference category. However, minor injury and property damage-only crashes did not show significantly higher odds of alcohol inference mismatch, aligning with expectations that fatal crashes undergo more detailed investigations and stricter reporting protocols.

Functional class of the road also played an important role. Crashes occurring on local roads had significantly lower log odds of alcohol inference mismatch (p = 0.041) compared to those on major roads, while crashes on collector roads and major roads did not exhibit significant differences. Additionally, log AADT (Average Annual Daily Traffic) showed a marginally significant effect (p = 0.122), indicating a slight decrease in the likelihood of alcohol inference mismatch with higher traffic volumes.

These findings underscore significant trends related to vehicle type, driver age, speed limits, road user types, light conditions, gender, location, and crash severity. Understanding these factors can guide targeted improvements in crash reporting systems, such as enhancing training programs for reporting officers and refining data collection protocols. These measures could contribute to more accurate and comprehensive crash data reporting, ultimately supporting better policymaking and road safety initiatives.

**Conclusion**

This study successfully bridges a critical research gap by developing a comprehensive framework for identifying crashes with AIM using NLP models. By focusing on inconsistencies between crash narratives and reported crash types, the study presents a systematic approach for detecting misclassified crashes, offering valuable insights into the accuracy of crash databases. The findings demonstrate the potential of applying NLP techniques, particularly the BERT model, to improve the reliability of crash-level data. The BERT model, trained on manually verified crash narratives, accurately classified alcohol and non-alcohol-related crashes and identified 2,767 underreported incidents, yielding an AIM rate of 24.03%. This rate is notably lower than the over 40% mismatch rate observed in previous studies (Miller, T., et al., 2012; Blincoe, L., et al., 2023). Furthermore, the regression analysis uncovered significant factors contributing to AIM, such as crash severity, vehicle type, driver age, and road user type. Notably, AIM was most prevalent in crashes with unknown or possible injuries and least common in fatal crashes.

This research not only identifies key factors that influence alcohol inference mismatch but also provides a practical framework for reducing database inaccuracies in crash reporting systems. By integrating this framework into crash data pipelines, agencies can improve the quality of crash data, ultimately aiding in more effective crash modeling and policymaking. Additionally, targeted revisions in law enforcement training programs—focusing on specific age groups, times of day, and vehicle types—could further reduce AIM. While this study significantly advances transportation research and law enforcement practices, future research could extend the application of NLP models to diverse crash reporting formats and types, while cross-verifying data with hospital records to enhance the overall reliability of crash-related information.

**Study Limitations**

One limitation of this study is the generalizability of the NLP models. These models are trained on a specific database format, potentially limiting their applicability across different jurisdictions. Another limitation is that NLP-based output must be validated by subject matter experts to ensure reliability and reduce misclassification risks.

## Acknowledgments

The crash data was provided by the Iowa Department of Transportation (DOT). The contents of this paper reflect the views of the authors and do not necessarily reflect the official views or policies of the sponsoring organization.

## Author contributions

The authors confirm contribution to the paper as follows: S. Bhagat conceived the study, carried out the analysis and drafted the paper. R. Kandiboina co-led the study and contributed to analysis, interpretation of results and drafting and finalizing the paper. I. Farabi Shihab helped in developing NLP mode. A. Sharma was the advisor who fine-tuned the study. S. Knickerbocker and N. Hawkins were committee members who fine-tuned the study. All authors reviewed the results and approved the final version of the manuscript.

## Conflict of interest:

The authors declare no conflicts of interest.